\documentclass{article}
\usepackage{spconf,amsmath,graphicx}
\usepackage{amsmath,amssymb,amsfonts}
\usepackage{algorithm}
\usepackage{algorithmic}
\usepackage{graphicx}
\usepackage{multicol}
\usepackage{tabularx}
\usepackage{float}
\usepackage{color, colortbl}
\usepackage{graphicx}
\usepackage{amsmath}
\usepackage{amssymb}
\usepackage{booktabs}
\usepackage{multirow}
\usepackage{multicol}
\usepackage{xcolor}
\usepackage{tabularx}
\usepackage{float}
\usepackage{hyperref}
\usepackage{setspace}

\definecolor{Cyan}{rgb}{0.88,1,1}
\definecolor{lightpurple}{RGB}{245,240,255}
\definecolor{lightgray}{RGB}{230,230,230}

\title{ViPo-MLLM: Visual-Pose Multimodal LLM for Gloss-Free Sign Language Translation}
\title{ViPo-MLLM: Visual-Pose Multimodal LLM for Gloss-Free Sign Language Translation}

\name{
Ahmed Abul Hasanaath\textsuperscript{1}, 
Bicheng Xu\textsuperscript{2}, 
Mir Rayat Imtiaz Hossain\textsuperscript{2}, 
Leonid Sigal\textsuperscript{2}, 
Hamzah Luqman\textsuperscript{1, *}\thanks{* Corresponding author: Hamzah Luqman, hluqman@kfupm.edu.sa}
}

\address{\textsuperscript{1}King Fahd University of Petroleum and Minerals, Dhahran, KSA\\
         \textsuperscript{2}University of British Columbia, Vancouver, Canada}

\begin{document}

\maketitle

\begin{abstract}
Gloss-free Sign Language Translation (SLT) translates sign language videos into spoken-language sentences without gloss annotations, avoiding costly labeling but requiring fine-grained modeling of hands, body, and facial cues. Existing methods often use single-modality or weakly fused features, limiting performance. We propose ViPo-MLLM, a framework that integrates spatio-temporal RGB and human pose features. Dedicated encoders model intra-modal dynamics and cross-modal attention captures long-range dependencies. The fused representation is conditioned with a structured prompt and processed by an LLM trained with contrastive and language modeling objectives. The proposed model was evaluated on the PHOENIX14T and CSL-Daily datasets and achieved new state-of-the-art results on both datasets. Moreover, the ViPo-MLLM model attained competitive performance compared to gloss-based recognition approaches, confirming the effectiveness of the proposed pose cues and cross-modal attention mechanisms. Code is available at \url{https://github.com/gufranSabri/ViPo-SLT}.
\end{abstract}


\section{Introduction}

Sign language conveys meaning through hand gestures, body movements, and facial expressions. Sign Language Translation (SLT) aims to bridge communication gaps by translating sign-language videos into spoken sentences. Despite recent progress, SLT remains challenging due to subtle temporal dynamics and the modality gap between visual input and natural language. Early SLT systems were largely gloss-based~\cite{camgoz2020sign, chen2024factorized}, relying on word-like annotations as intermediates. Although effective, gloss annotation is costly, limits scalability, and can produce rigid sentence structures. These limitations have motivated gloss-free approaches~\cite{zhou2023gloss, kim2025leveraging} that directly translate videos into fluent spoken sentences.

Large Language Models (LLMs)~\cite{gong2024llms, jang2025lost, wong2024sign2gpt} enhance gloss-free SLT, but their performance hinges on how signing is represented. Some methods use separate visual encoders~\cite{gong2024llms, jang2025lost}, while others employ multimodal encoders~\cite{kim2025leveraging}. These representations often require careful alignment and may fail to capture fine-grained articulatory details critical for accurate translation.

A central challenge is capturing linguistically relevant components such ashand gestures, facial cues, while remaining learnable and alignable with language. RGB video encodes global appearance and context, while pose keypoints provide a compact, structured description of articulator motion. Combining these cues is non-trivial: naive fusion often fails to align pose and visual evidence or capture cross-modal dependencies.  

We propose \textbf{V}isual-\textbf{P}ose \textbf{M}ultimodal \textbf{LLM} (ViPo-MLLM), a cross-modal attention framework for gloss-free SLT. Dedicated encoders extract RGB and pose features, which are fused via time-aware cross-modal attention and decoded into spoken sentences using a transformer, trained end-to-end without gloss supervision.


Our contributions are fourfold: (1) we propose ViPo-MLLM, a framework that integrates spatio-temporal RGB features and pose cues through cross-modal attention; (2) we introduce an explicit video-pose fusion strategy that models \emph{when} and \emph{how} pose information influences translation; (3) we conduct extensive experimental analyses that reveal the complementary roles of pose information and spatio-temporal visual cues in gloss-free SLT; and (4) we achieve a new state-of-the-art performance on the PHOENIX14T and CSL-Daily datasets.

\section{Literature Review}
\label{sec:lit}

\begin{figure*}[t]
    \centering
    \includegraphics[width=0.8\textwidth]{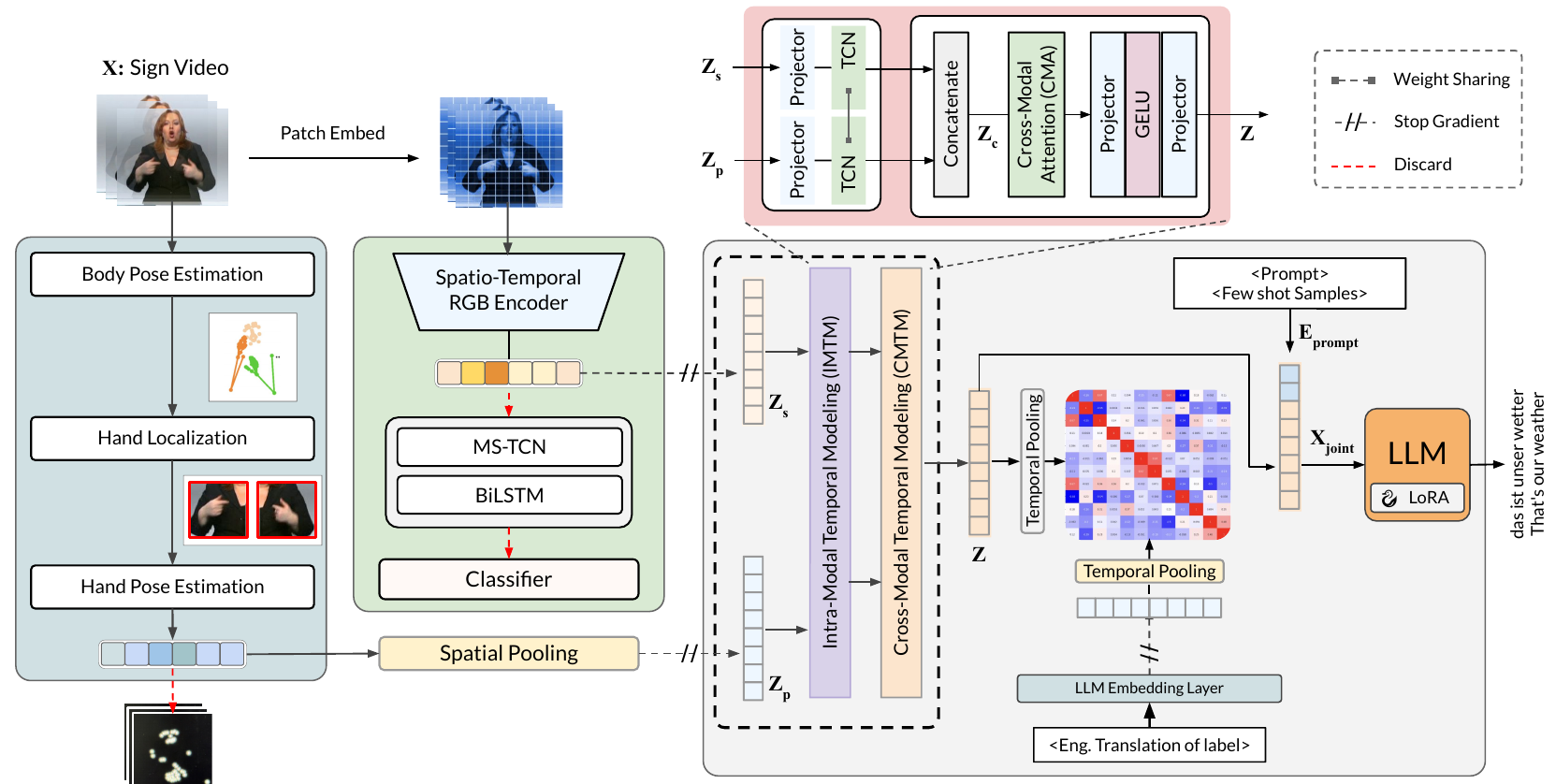}
    \caption{\textbf{Pipeline of the ViPo-MLLM framework.} The spatio-temporal and pose-based features are first extracted from a sign language video through a USTM-based encoder and OpenPose, respectively. These complementary features are then fused via the Intra-Modal and Cross-Modal Temporal Modeling blocks to capture both local and global dependencies. The resulting multimodal representations are combined with a structured prompt and fed into an LLM, which generates gloss-free translations.}
    \label{fig:pipeline}
\end{figure*}

\vspace{1mm}\noindent\textbf{Gloss-Free SLT. }
Gloss-free SLT translates sign language videos directly into spoken sentences without glosses, but often underperforms gloss-based methods. Prior work addresses this through \emph{temporal and semantic modeling} and \emph{cross-modal alignment}. TSPNet \cite{li2020tspnet} models hierarchical temporal semantics, while CSGCR \cite{zhao2021conditional} uses cross-modal re-ranking for translation selection. ConSLT \cite{fu2023token}, GASLT \cite{yin2023gloss}, and VAP \cite{jiao2024visual} further improve visual-text alignment using contrastive learning, attention constraints, and gloss-like supervision.

Recent methods leverage \emph{large pretrained models} to inject linguistic knowledge. GFSLT-VLP \cite{zhou2023gloss} aligns video features with language via visual-language pretraining. Sign2GPT \cite{wong2024sign2gpt} and FLa-LLM \cite{chen2024factorized} use pretrained models with adapters or factorized training to improve visual representations without gloss supervision. SignLLM \cite{gong2024llms} maps sign videos to discrete tokens aligned with LLMs. SpaMo \cite{hwang2024efficient} emphasizes spatial and motion cues, and MMSLT \cite{kim2025leveraging} exploits multimodal LLMs to generate detailed textual descriptions aligned with video.

\vspace{1mm}\noindent\textbf{Multimodal SLT. }
Early SLT mainly relied on \emph{unimodal visual modeling}, demonstrating end-to-end RGB-based translation \cite{camgoz2020sign, kayo2020better}, later refined for efficiency and alignment \cite{jang2025lost2}. Multimodal SLT exploits multiple channels to capture richer semantics. \emph{Multi-cue modeling} fuses optical flow, skeletal, handshape, and mouthing features \cite{papadimitriou2020multimodal, camgoz2020multi}, while STMC \cite{zhou2021spatial} jointly models spatial and temporal cues. \emph{Structured interaction} uses graph-based encodings (MSeqGraph \cite{tang2021graph}) and skeleton-aware scaling (SANet \cite{gan2021skeleton}), while multi-stream encoders like TwoStream-SLT \cite{chen2022two} reduce redundancy. Recent work leverages \emph{large-scale pretraining and multimodal LLMs}. MMSLT \cite{kim2025leveraging} aligns video features with textual descriptions.

\section{Methodology}
\label{sec:methodology}

\subsection{Framework Overview}
Figure~\ref{fig:pipeline} shows the ViPo-MLLM architecture. Spatio-temporal and pose-based features are extracted from a sign video using a USTM encoder~\cite{hasanaath2025ustmunifiedspatialtemporal} and OpenPose~\cite{cao2019openpose}, respectively. These features are fused via Intra-Modal and Cross-Modal Temporal blocks to capture local and global dependencies. The fused representation, combined with a structured prompt, is fed to an LLM for gloss-free translation. The model is trained with contrastive and cross-entropy losses to align visual-pose cues with text. Below, we describe sign feature extraction, multimodal fusion, prompt design, and training objectives.

\subsection{Sign Feature Extraction}

Given a sign video $X = \{x_t\}_{t=1}^{T}$, the goal is to extract visual representations that capture both spatio-temporal cues and fine-grained articulatory dynamics. ViPo-MLLM uses two complementary streams: spatio-temporal RGB features $\mathbf{Z}_{\text{s}}$ and pose-based features $\mathbf{Z}_{\text{p}}$.

\vspace{2mm}\noindent\textbf{Spatio-Temporal Features.}
We extract spatio-temporal features using the USTM framework, which employs a hierarchical vision encoder with stacked spatio-temporal modules combining Swin Transformer stages and lightweight temporal adapters. We use pretrained recognition weights and keep the encoder frozen, discarding late-stage components (MS-TCN, BiLSTM, classification heads) to avoid task-specific temporal bias. This produces frame-level spatio-temporal representations $\mathbf{Z}{\text{s}} \in \mathbb{R}^{T \times D{s}}$, where $D_s$ is the embedding dimension.

\vspace{2mm}\noindent\textbf{Pose Features.}
Pose-based features are extracted using OpenPose, which estimates full-body and fine-grained hand poses. Instead of final heatmaps, we use intermediate features before heatmap generation, applying spatial average pooling per frame to obtain fixed-dimensional vectors. Aggregating over time yields $\mathbf{Z}_{p} \in \mathbb{R}^{T \times D_{p}}$, where $D_p$ is the embedding dimension.

\vspace{2mm}\noindent\textbf{Multimodal Feature Fusion.}
To integrate spatio-temporal and pose features, we propose a multimodal fusion network modeling intra- and cross-modal temporal dependencies. 

The \emph{Intra-Modal Temporal Modeling} (IMTM) block projects $\mathbf{Z}_{\text{s}}$ and $\mathbf{Z}_{\text{p}}$ into a shared $D$-dimensional space using a linear layer, then captures local temporal dynamics via a 2-layer 1D Temporal Convolutional Network (TCN) \cite{lea2016temporalconvolutionalnetworksaction} with kernel size 5 and max-pooling. The same TCN is shared across modalities to enforce consistent temporal patterns.

The \emph{Cross-Modal Temporal Modeling} (CMTM) block concatenates IMTM outputs along time to form $\mathbf{Z}_{c} \in \mathbb{R}^{2T' \times D}$, and passes them through a 4-layer \emph{Cross-Modal Attention} (CMA) module, enabling global temporal modeling and inter-modal interactions.
Finally, an MLP projects the fused representation to the LLM embedding dimension $\mathbf{Z} \in \mathbb{R}^{2T' \times D_L}$.


\subsection{Language Modeling}
\vspace{1.5mm}\noindent\textbf{Prompt Construction.}
To condition the LLM on multimodal features, we construct a structured prompt combining textual instructions and few-shot examples. The instruction template ``Translate the given sentence into \texttt{<language>}'' is tokenized and embedded, and three few-shot translations (Spanish, French, English $\rightarrow$ \texttt{<language>}) are similarly embedded. 
Concatenating the instruction and examples yields the complete prompt embedding:

\[
\mathbf{E}_{\text{prompt}} = \mathrm{Concat}(\mathbf{E}_{\text{instr}}, \mathbf{E}_{\text{ex}}) \in \mathbb{R}^{L_p \times D}, \quad L_p = L_{\text{instr}} + L_{\text{ex}}
\]

The multimodal features $\mathbf{Z}$ are then concatenated with $\mathbf{E}_{\text{prompt}}$ along the temporal dimension to form the LLM input:
\[
\mathbf{X}_{\text{joint}} = \mathrm{Concat}(\mathbf{E}_{\text{prompt}}, \mathbf{Z}) \in \mathbb{R}^{L_p + 2T' \times D},
\]
integrating textual guidance with visual-pose context for translation.

\vspace{2mm}\noindent\textbf{Learning Objective.}
To align multimodal visual-pose features with text, we optimize a combination of contrastive and cross-entropy losses. Temporal pooling of $\mathbf{Z}$ yields a global video representation $\mathbf{z} \in \mathbb{R}^{D}$. English translations of the ground-truth sentences are tokenized and embedded via the frozen LLM to produce pseudo-label embeddings $\mathbf{Y}_{\text{pseudo}} \in \mathbb{R}^{L \times D}$, pooled to a global representation $\mathbf{y} \in \mathbb{R}^{D}$. Contrastive learning between $\mathbf{z}$ and $\mathbf{y}$ gives:
\[
\mathcal{L}_{\text{cont}} = - \log \frac{\exp(\mathbf{z} \cdot \mathbf{y} / \tau)}{\sum_{j} \exp(\mathbf{z} \cdot \mathbf{y}_j / \tau)},
\]
with learnable temperature $\tau$. Simultaneously, the joint input $\mathbf{X}_{\text{joint}}$ is fed to the LLM to compute the cross-entropy loss:
\[
\mathcal{L}_{\text{CE}} = \mathrm{CrossEntropy}(\mathrm{LLM}(\mathbf{X}_{\text{joint}}), \mathbf{Y}_{\text{target}}).
\]
The total loss is: $\mathcal{L} = \mathcal{L}_{\text{cont}} + \mathcal{L}_{\text{CE}}$, aligning video and textual representations while predicting translations.

\section{Experiments}
\label{sec:results}

\subsection{Experimental Setup}

\vspace{1mm}\noindent\textbf{Datasets.}
We evaluate on two benchmark SLT datasets: PHOENIX14T \cite{phoenix14t} and CSL-Daily \cite{csldaily}. PHOENIX14T contains 8,247 German sign language videos with 1,085 glosses (avg. 7.7 per sentence), split into 7,096 training, 519 validation, and 642 test samples. CSL-Daily has 20,654 Chinese sign language videos covering ~2,000 glosses (avg. 7.2 per sentence), split into 18,401 training, 1,077 validation, and 1,176 test samples.

\begin{table*}[ht]
\centering
\footnotesize
\setlength{\tabcolsep}{2.7pt}
\caption{Comparison with gloss-based, weakly supervised, and gloss-free methods on Phoenix14T and CSL-Daily.}
\label{tab:sota_slt}
\begin{tabular}{lccccc|ccccc}
\toprule
\multirow{2}{*}{\textbf{Model}} &
\multicolumn{5}{c}{\textbf{Phoenix14T}} &
\multicolumn{5}{c}{\textbf{CSL-Daily}} \\ 
& BLEU1 & BLEU2 & BLEU3 & BLEU4 & ROUGE-L
& BLEU1 & BLEU2 & BLEU3 & BLEU4 & ROUGE-L \\

\midrule
\multicolumn{11}{c}{\textbf{Gloss-based}} 
\\ \midrule

SLRT \cite{camgoz2020sign} & 46.61 & 33.73 & 26.19 & 21.32 & - & 37.38 & 24.36 & 16.55 & 11.79 & 36.74 \\
STN-SLT \cite{voskou2021stochastic} & 48.61 & 35.97 & 28.37 & 23.65 & - & - & - & - & - & - \\
STMC-T \cite{zhou2021spatial} & 46.98 & 36.09 & 28.70 & 23.65 & 46.65 & - & - & - & - & - \\
TS-SLT \cite{chen2022two} & 54.90 & 42.43 & 34.46 & 28.95 & 53.48 & 55.44 & 42.59 & 32.87 & 25.79 & 55.72 \\
MSKA \cite{guan2025mska} & 54.79 & 42.42 & 34.49 & 29.03 & 53.54 & 56.37 & 42.80 & 32.78 & 25.52 & 54.04 \\

\midrule
\multicolumn{11}{c}{\textbf{Weakly Supervised Gloss-free}} \\ \midrule

TSPNet \cite{li2020tspnet} & - & - & - & - & - & 17.09 & 8.98 & 5.07 & 2.97 & 18.38 \\
GASLT \cite{yin2023gloss} & - & - & - & - & - & 19.00 & 9.94 & 5.98 & 4.07 & 20.35 \\
ConSLT \cite{fu2023token} & - & - & - & 21.11 & 47.74 & - & - & - & 14.53 & 40.98 \\
VAP \cite{jiao2024visual} & 52.78 & - & - & 26.62 & 51.47 & 49.99 & - & - & 20.85 & 48.56 \\

CMDA-SLT \cite{ye2305cross} & - & - & - & 25.36 & 49.87 & - & - & - & 21.58 & 49.34 \\

\midrule
\multicolumn{11}{c}{\textbf{Gloss-free}} \\ \midrule

CSGCR \cite{zhao2021conditional} & 36.71 & 25.40 & 18.86 & 15.18 & 38.85 & - & - & - & - & - \\
GFSLT-VLP \cite{zhou2023gloss} & 43.71 & 33.18 & 26.11 & 21.44 & 42.49 & 39.37 & 24.93 & 16.26 & 11.00 & 36.44 \\
Sign2GPT \cite{wong2024sign2gpt} & 49.54 & 35.96 & 28.83 & 22.52 & 48.90 & 41.75 & 28.73 & 20.60 & 15.40 & 42.36 \\
FLa-LLM \cite{chen2024factorized} & 46.29 & 35.33 & 28.03 & 23.09 & 45.27 & 37.13 & 25.12 & 18.38 & 14.20 & 37.25 \\
SignLLM \cite{gong2024llms} & 45.21 & 34.78 & 28.05 & 23.40 & 44.49 & 39.55 & 28.13 & 20.07 & 15.75 & 39.91 \\
SpaMo \cite{hwang2024efficient} & \underline{49.80} & 37.30 & 29.50 & 24.30 & 46.60 & 48.90 & \underline{36.90} & 26.78 & 20.55 & 47.46 \\
MMSLT \cite{kim2025leveraging} & 48.90 & \underline{38.10} & \underline{30.80} & \underline{25.70} & \underline{48.00} & \underline{49.87} & 36.37 & \underline{27.29} & \underline{21.11} & \underline{48.92} \\

\midrule
\rowcolor{lightgray}
\textbf{ViPo-MLLM (Ours)} & \textbf{50.97} & \textbf{40.00} & \textbf{32.36} & \textbf{27.10} & \textbf{51.50}
              & \textbf{56.13} & \textbf{42.45} & \textbf{32.74} & \textbf{25.85} & \textbf{54.53} \\
\bottomrule
\end{tabular}
\end{table*}

\vspace{2mm}\noindent\textbf{Implementation Details.}
We use mT0-XL~\cite{muennighoff2023crosslingual} as the backbone LLM with LoRA~\cite{hu2021loralowrankadaptationlarge} (rank $r=16$) for efficient fine-tuning. Training has two stages: 100 warm-up epochs optimizing only the contrastive objective, followed by 50 epochs of joint contrastive and language modeling optimization. A Reduce-on-Plateau scheduler (patience 5, gamma 0.2) is applied. Experiments run on a single NVIDIA H100 GPU (80GB VRAM), and evaluation uses BLEU-1–4 and ROUGE-L metrics.

\subsection{Comparison with State-of-the-Art}
Table \ref{tab:sota_slt} compares ViPo-MLLM with \emph{gloss-based}, \emph{weakly supervised gloss-free}, and \emph{fully gloss-free} SLT methods. Our main comparison is with weakly supervised gloss-free approaches, as ViPo-MLLM uses a pretrained RGB encoder but remains gloss-free during SLT training. We also include fully gloss-free and gloss-based methods for context.

On PHOENIX14T, ViPo-MLLM achieves BLEU-4 27.10 and ROUGE-L 51.50, outperforming weakly supervised and fully gloss-free methods such as VAP \cite{jiao2024visual}, ConSLT \cite{fu2023token}, SpaMo \cite{hwang2024efficient}, and MMSLT \cite{kim2025leveraging}. Compared to gloss-based approaches, it remains competitive despite not using gloss annotations. On CSL-Daily, our framework attains BLEU-4 25.85 and ROUGE-L 54.53, achieving SOTA among gloss-free methods and comparable performance to gloss-based models. This demonstrates that gloss-free training can match or exceed gloss-supervised approaches.

\subsection{Ablation Studies}
We conduct ablation studies on the PHOENIX14T dataset to systematically analyze the contribution of each component in the ViPo-MLLM framework. Specifically, we evaluate (i) the impact of the main architectural components, (ii) the effect of varying the depth of the CMA module, (iii) the influence of different LLM backbones, and (iv) the effect of prompt formulation. All ablations are performed under identical training settings and evaluated using BLEU1-BLEU4 and ROUGE-L.

\begin{table}[ht]
\centering
\scriptsize
\setlength{\tabcolsep}{4pt}
\caption{Ablation study on the contribution of RGB, pose, and contrastive learning (Cont.) components.}
\label{tab:ablation_components}
\begin{tabular}{ccc|ccccc}
\toprule
\textbf{RGB} & \textbf{Pose} & \textbf{Cont.} & \textbf{BLEU1} & \textbf{BLEU2} & \textbf{BLEU3} & \textbf{BLEU4} & \textbf{ROUGE-L} \\
\midrule
$\checkmark$ &      &      & 48.16 & 37.71 & 30.53 & 25.54 & 48.36 \\
$\checkmark$ &      & $\checkmark$ & 49.60 & 38.55 & 30.95 & 25.64 & 49.27 \\
     & $\checkmark$ &      & 32.31 & 21.99 & 16.07 & 12.88 & 30.13 \\
     & $\checkmark$ & $\checkmark$ & 34.56 & 24.20 & 18.30 & 14.62 & 33.52 \\
$\checkmark$ & $\checkmark$ &      & 49.02 & 38.72 & 31.58 & 26.61 & 49.57 \\
$\checkmark$ & $\checkmark$ & $\checkmark$ & \textbf{50.97} & \textbf{40.00} & \textbf{32.36} & \textbf{27.10} & \textbf{51.50} \\
\bottomrule
\end{tabular}
\end{table}

\vspace{2mm}\noindent\textbf{Effect of Main Components.}
Table~\ref{tab:ablation_components} compares RGB and pose modalities. Using RGB alone provides a strong baseline (BLEU-4: 25.54, ROUGE-L: 48.36), while pose alone performs worse (BLEU-4: 12.88, ROUGE-L: 30.13), showing limited expressiveness. Contrastive supervision improves both: RGB with contrastive learning achieves 25.64 BLEU-4 and 49.27 ROUGE-L, and pose improves to 14.62 BLEU-4 and 33.52 ROUGE-L, indicating enhanced semantic alignment.

Combining RGB and pose without contrastive learning boosts performance (BLEU-4: 26.61, ROUGE-L: 49.57), confirming their complementarity. Applying contrastive learning to the combined features yields the best results, highlighting the importance of joint multimodal modeling and semantic alignment for gloss-free SLT.

\begin{table}[ht]
\centering
\scriptsize
\setlength{\tabcolsep}{7pt}
\caption{Ablation study on the depth of CMA.}
\label{tab:cma_stages}
\begin{tabular}{lccccc}
\toprule
\textbf{CMA Depth} & \textbf{BLEU1} & \textbf{BLEU2} & \textbf{BLEU3} & \textbf{BLEU4} & \textbf{ROUGE-L} \\
\midrule
0 & 49.88 & 38.58 & 30.91 & 25.67 & 48.86 \\
1 & 50.28 & 39.41 & 31.97 & 26.75 & 50.76 \\
2 & 50.32 & 39.35 & 31.81 & 26.53 & 51.01 \\
4 & \textbf{50.97} & \textbf{40.00} & \textbf{32.36} & \textbf{27.10} & \textbf{51.50} \\
8 & 49.46 & 38.72 & 31.25 & 26.08 & 50.33 \\
\bottomrule
\end{tabular}
\end{table}

\vspace{2mm}\noindent\textbf{Effect of CMA Depth.}
Table~\ref{tab:cma_stages} shows ViPo-MLLM performance across CMA depths. Without CMA (depth 0, similar to SpaMo~\cite{hwang2024efficient}), BLEU-4 is 25.67 and ROUGE-L is 48.86. Adding CMA improves results at depths 1 and 2, with performance peaking at depth 4 (BLEU-4: 27.10, ROUGE-L: 51.50), indicating an optimal trade-off between capacity and generalization. Further increasing depth to 8 degrades performance, likely due to over-smoothing or overfitting.

\begin{table}[ht]
\centering
\scriptsize
\setlength{\tabcolsep}{3pt}
\caption{Ablation study on the impact of the LLM, comparing model \textit{size} (mT0 family) and model \textit{choice} across different pretrained LLMs.}
\label{tab:lora_vs_fullft}
\begin{tabular}{l|ccccc}
\toprule
\textbf{Model} & \textbf{BLEU1} & \textbf{BLEU2} & \textbf{BLEU3} & \textbf{BLEU4} & \textbf{ROUGE-L} \\
\midrule
bigscience/mT0-base \cite{muennighoff2023crosslingual}  & 46.62 & 35.98 & 28.71 & 23.82 & 48.37 \\
bigscience/mT0-large \cite{muennighoff2023crosslingual} & 49.74 & 38.72 & 31.34 & 26.27 & 50.17 \\
bigscience/mT0-xl \cite{muennighoff2023crosslingual}    & \textbf{50.97} & \textbf{40.00} & \textbf{32.36} & \textbf{27.10} & \textbf{51.50} \\
\midrule
google/flan-t5-xl \cite{chung2022scalinginstructionfinetunedlanguagemodels}        & 50.40 & 39.72 & 32.28 & 27.05 & 51.32 \\
facebook/mbart-large-50 \cite{liu2020multilingualdenoisingpretrainingneural} & 45.11 & 34.39 & 27.13 & 22.36 & 45.67 \\
\bottomrule
\end{tabular}
\end{table}

\vspace{1mm}\noindent\textbf{Effect of LLM Choice.}
Table~\ref{tab:lora_vs_fullft} shows the impact of LLM size and type. Within the mT0 family, larger models improve performance, with mT0-xl achieving the best scores (BLEU-4: 27.10, ROUGE-L: 51.50). Flan-T5-XL performs similarly, but lacks non-Latin support for CSL-Daily. Smaller models like mbart-large-50 lag behind (BLEU-4: 22.64, ROUGE-L: 45.67), highlighting the importance of model capacity.

\begin{table}[ht]
\centering
\scriptsize
\setlength{\tabcolsep}{5pt}
\caption{Ablation study on the effect of prompt structure.}
\label{tab:prompt_structure}
\begin{tabular}{l|ccccc}
\toprule
\textbf{Prompt Type} & \textbf{BLEU1} & \textbf{BLEU2} & \textbf{BLEU3} & \textbf{BLEU4} & \textbf{ROUGE-L} \\
\midrule
None & 50.55 & 39.74 & 32.16 & 26.89 & 50.87 \\
INSTRUCTIVE & \textbf{50.97} & \textbf{40.00} & \textbf{32.36} & \textbf{27.10} & \textbf{51.50} \\
INTERROGATIVE & 48.99 & 38.20 & 30.81 & 25.60 & 49.49 \\
ROLE PLAYING & 43.90 & 33.15 & 26.14 & 21.42 & 43.23 \\
\bottomrule
\end{tabular}
\end{table}

\vspace{2mm}\noindent\textbf{Effect of Prompt Structure.}
Different prompt formulations can significantly influence the generation behavior of an LLM. Following the prompt types proposed in~\cite{hasanaath2025arareasoner}, we evaluate several prompting strategies. Table~\ref{tab:prompt_structure} analyzes the impact of different prompt formulations. Using no prompt already yields strong performance (BLEU-4: 26.89, ROUGE-L: 50.87), indicating that the visual encoder and LLM alignment are effective even without explicit instruction tuning. However, the instructive prompt achieves the best results across all metrics (BLEU-4: 27.10, ROUGE-L: 51.50), suggesting that direct task-oriented guidance helps the LLM better interpret visual sign representations. In contrast, interrogative prompts reduce performance (BLEU-4: 25.60, ROUGE-L: 49.49), likely because framing the task as a question introduces unnecessary ambiguity. Role-playing prompts perform substantially worse (BLEU-4: 21.42, ROUGE-L: 43.23), indicating that conversational or persona-based prompting distracts the model from the translation objective.
\section{Conclusion}
\label{sec:conclusion}

We proposed ViPo-MLLM, an SLT framework that fuses spatio-temporal RGB features with pose representations via cross-modal temporal attention. By modeling intra-modal dynamics and cross-modal dependencies, and leveraging structured prompting with contrastive alignment, our method captures fine-grained motion and long-range temporal patterns crucial for translation. Experiments on PHOENIX14T and CSL-Daily show SOTA performance among gloss-free methods, remaining competitive with gloss-based pipelines. Ablations confirm the complementary value of pose and visual features and the effectiveness of cross-modal attention. Future work will explore additional modalities, lightweight pose representations, and improved temporal alignment for scalable, real-time SLT.

\section{Acknowledgements}
\label{sec:ack}

The authors would like to acknowledge the support provided by King Fahd University of Petroleum \& Minerals (KFUPM) for funding this work through project number ISP24226.

\bibliographystyle{IEEEbib}
\bibliography{strings,refs}

\end{document}